\documentclass[preprint]{elsarticle}

\usepackage{lineno,hyperref}
\usepackage{amssymb}
\usepackage{graphicx}
\usepackage{multirow}
\usepackage{caption}
\usepackage{subfigure}
\usepackage[linesnumbered,ruled,vlined]{algorithm2e}
\usepackage{algorithmic}
\SetKwInput{KwInput}{Input}                
\SetKwInput{KwOutput}{Output}              
\usepackage{amsmath}
\modulolinenumbers[5]
\usepackage{setspace}

\journal{Knowledge-Based Systems}
\begin{document}

\begin{frontmatter}

\title{Online Active Proposal Set Generation for Weakly Supervised Object Detection}

\author[addr1]{Ruibing Jin}
\author[addr2]{Guosheng Lin\corref{CorrAuthor}}
\author[addr1]{Changyun Wen\corref{CorrAuthor}}

\cortext[CorrAuthor]{Corresponding author}
\ead{gslin@ntu.edu.sg (Guosheng Lin) and ecywen@ntu.edu.sg}

\address[addr1]{School of Electrical and Electronic Engineering, Nanyang Technological University (NTU), Singapore 639798}
\address[addr2]{School of Computer Science and Engineering, Nanyang Technological University (NTU), Singapore 639798}

\begin{abstract}
To reduce the manpower consumption on box-level annotations, many weakly supervised object detection methods which only require image-level annotations, have been proposed recently. 
The training process in these methods is formulated into two steps. 
They firstly train a neural network under weak supervision to generate pseudo ground truths (PGTs). Then, these PGTs are used to train another network under full supervision. 
Compared with fully supervised methods, the training process in weakly supervised methods becomes more complex and time-consuming. 
Furthermore, overwhelming negative proposals are involved at the first step. This is neglected by most methods, which makes the training network biased towards to negative proposals and thus degrades the quality of the PGTs, limiting the training network performance at the second step.
Online proposal sampling is an intuitive solution to these issues. However, lacking of adequate labeling, a simple online proposal sampling may make the training network stuck into local minima. To solve this problem, we propose an \textbf{O}nline Active \textbf{P}roposal Set \textbf{G}eneration (OPG) algorithm. Our OPG algorithm consists of two parts: Dynamic Proposal Constraint (DPC) and Proposal Partition (PP). DPC is proposed to dynamically determine different proposal sampling strategy according to the current training state. PP is used to score each proposal, part proposals into different sets and generate an active proposal set for the network optimization. Through experiments, our proposed OPG shows consistent and significant improvement on both datasets PASCAL VOC 2007 and 2012, yielding comparable performance to the state-of-the-art results.
\end{abstract}


\end{frontmatter}

\section{Introduction}

Objection detection \cite{girshick2015fast,ren2015faster,redmon2016you,liu2016ssd,li2020weak,perez2020object,wei2020incremental,wang2019detection} achieves significant advance benefited from the remarkable progress of Convolutional Neural Network (CNN) \cite{simonyan2014very,he2016deep,szegedy2015going}. However, training a fully supervised object detection network often requires  numerous training images with expensive bounding box annotations.
To alleviate this issue, weakly supervised object detection which only requires image-level
annotations, receives much attention recently, where the labels can be easily obtained from image tags. 

The training process in existing weakly supervised object detection methods consists of two steps. In the first step, inspired by the proposal-based detection network in the fully supervised setting, many works \cite{tang2017multiple,tang2018weakly,wei2018ts2c,wan2019c,shen2019cyclic}  formulate the weakly supervised object detection task as a Multiple Instance Learning (MIL) problem, where each image is regarded as a bag and the proposals are regarded as instances. The network is trained to predict the proposal scores. At the second step, pseudo ground truths (PGTs) are produced by using the trained network at the first step. These PGTs are used as box-level annotations and another network is trained under full supervision.

Compared with the training process under full supervision, this two-step training process is complex and time-consuming. Moreover, since the original proposal sets are dominated by overwhelming negative proposals, all proposals used for training at the first step may causes the training network to be biased towards negative proposals \cite{ren2015faster}. This further affects the quality of the PGTs and limits the performance of the network trained at the second step.

The issue arisen by these overwhelming negative proposals widely exists in detection methods. In fully supervised approaches, like Fast R-CNN\cite{girshick2015fast}, 
Faster R-CNN \cite{ren2015faster} and SSD \cite{liu2016ssd}, they use box-level annotations filter out the redundant negative proposals, maintaining the ration between positive and negative proposals at 1:1 or 1:3, listed in Table \ref{table:fb_ratio}. Without box-level annotations, weakly supervised methods use all proposals during training at the first step, resulting in the ratio around 1:100. This unreasonable ratio may have an influence on the performance of weakly supervised object detection methods. However, this problem is neglected by most weakly supervised methods \cite{bilen2016weakly,tang2017multiple,tang2018weakly,zhang2018zigzag,kantorov2016contextlocnet,wei2018ts2c,wan2019c,shen2019cyclic,singh2019you}. To show its importance and encourage people to focus on this issue, a series of experiments are firstly conducted to show the ratio influence on the detection accuracy in Section \ref{sec3}. 

\begin{table}[htbp]
	\caption{The ratio between positive and negative proposals in training proposal sets. WSOD is short for weakly supervised object detection methods.}
	\begin{center}
		\begin{tabular}{c|c|c|c}
			\hline
			\multicolumn{2}{c|}{\multirow{2}{*}{Methods}}    & Supervised     & \multirow{2}{*}{Ratio}\\
			\multicolumn{2}{c|}{}                            & Setting          &                       \\
			\hline
			\multicolumn{2}{c|}{Fast R-CNN} &  Full                 &   1:3  \\
			\hline
			\multirow{2}{*}{Faster R-CNN} &  RPN          & Full          & 1:1  \\
			\cline{2-4}
			&  R-CNN        &	Full          & 1:3  \\ 						
			\hline
			\multicolumn{2}{c|}{SSD}                      & Full          & 1:3 \\
			\hline
			\multicolumn{2}{c|}{WSOD}     & Weak & $\approx$ \textbf{1:100} \\
			\hline
		\end{tabular}
	\end{center}
	\label{table:fb_ratio}
\end{table}

Then, we try to alleviate the affect caused by overwhelming negative proposals. Proposal sampling is an effective method. In full supervision, Focal loss \cite{lin2018focal} and OHEM \cite{shrivastava2016training} are two well-known approaches to mine hard proposals for training. However, these methods are proposed based on box-level annotations. Since weakly supervised methods lack box-level annotations, Focal loss and OHEM are not applicable to weakly supervised methods. 

\begin{figure}[htbp]
	\centering
	{\includegraphics[width=.6\textwidth]{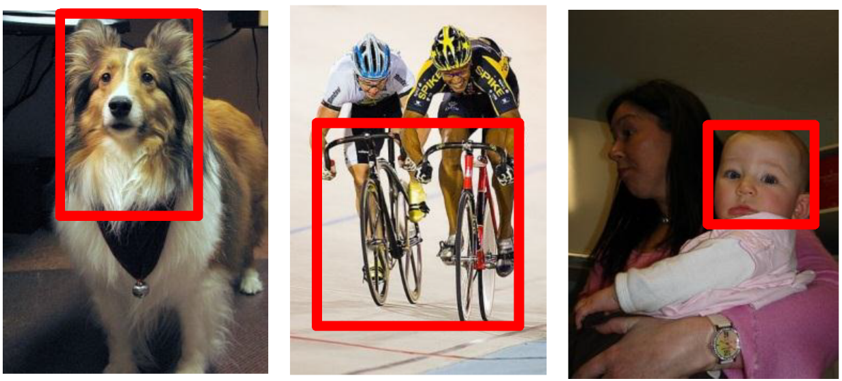}
	}
	
	\caption{False detection results in weakly supervised detection methods. Pseudo ground truths based on these inaccurate results may degrade the training network performance.
	}
	\label{fig:local}
\end{figure}

In this paper, we aim to propose an algorithm which combines the two-step training into one-step and mitigates the issue arisen by the redundant negative proposals. An intuitive method may produce the PGTs in an online manner and use these PGTs to sample proposals. However, under weak supervision, the training model especially for the first several epochs, is prone to get stuck into local minima, regarding object parts as whole objects. As illustrated in Fig. \ref{fig:local}, weakly supervised methods tend to focus on discriminative regions. The head of the dog is recognized as an object, while other parts of it are neglected. Two bicycles are detected as a whole object and only the head of the baby is detected. The PGTs produced based on this inaccurate detection results may destroy the network training and degenerate the training network performance.

To address this issue, we propose an \textbf{O}nline Active \textbf{P}roposal Set \textbf{G}eneration (OPG) algorithm. Our OPG algorithm is composed of two parts: Dynamic Proposal Constraint (DPC) and Proposal Partition (PP). Our OPG aims to produce PGTs in an online manner. During the training process, the training network performance is not stable. The PGTs produced by the training network, especially for first several epochs are not accurate, even destroying the training process. To solve this problem, DPC is proposed to dynamically determine the proposal sample strategy according to the training state. Our DPC controls the number of filtered proposal during the training process. It suppress the effectiveness of proposal sampling at the beginning of the training process, while enhances its effectiveness at the end of the training process. PP is proposed to score proposals, divide proposals into positive or negative sets and generate an active proposal set. During the training process, the training network is optimized based on our produced active proposal set.  

\begin{figure*}[htbp]
	\centering
	{\includegraphics[width=\textwidth]{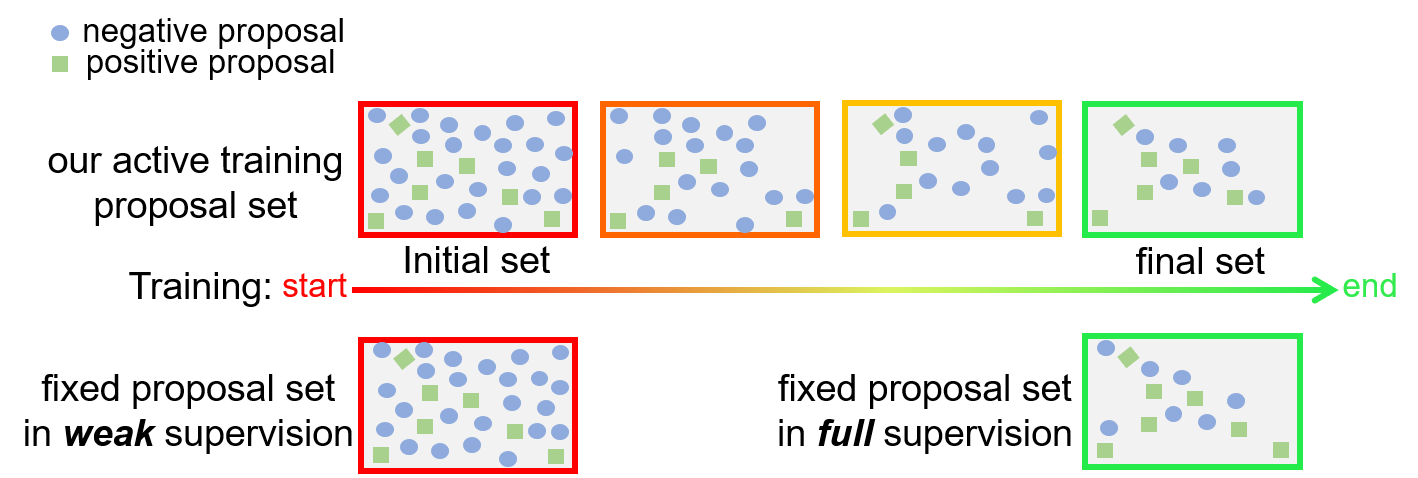}
	}
	\caption{Diagram of our expected active training proposal set from our \textbf{O}nline Active \textbf{P}roposal Set \textbf{G}eneration (OPG) in training. }
	\label{fig:diagram}
\end{figure*} 

To present our OPG clearly, a diagram for our expected active proposal sets during the training process is illustrated in Fig.\ref{fig:diagram}, where circle points and rectangles points indicate negative and positive proposals, respectively. Our active training proposal set is initialized with the original training proposal set which
is the same as the training proposal set in existing weakly supervised object detection methods. During the training procedure, our OPG
utilizes the predictions from the training model to online generate an active training proposal set.
In this process, the issue caused by overwhelming negative samples can be gradually alleviated. Finally, the positive and negative samples in our generated active set is balanced, which is similar to that of fully supervised methods. Experimental results show that our proposed OPG is able to
effectively improve the baseline and yield a SOTA performance.

Recently, some works\cite{tang2018weakly,singh2019you} propose to generate high-quality proposals with various region proposal networks (RPNs). These methods generally requires a complex training procedures or an external dataset, while our proposed OPG does not need any extra operations. Our OPG algorithm can be easily applied to other methods, improving their performances.

Overall, our contribution can be summarized as follows: 1) To emphasize the importance of the ratio between positive and negative proposals in training, we conduct a series of analysis experiments on the influence caused by this ratio on detection performance. 2) A Dynamic Proposal Constraint (DPC) is proposed. Our DPC dynamically chooses proposal sampling strategy, which ensures that the network is trained in a stable way and prevents the training network from falling into a local minima. 3) In our OPG algorithm, a Proposal Partition (PP) component is proposed to score proposal, part proposals into positive, negative and risk sets, and generate the active proposal set for training. 4) Extensive experiments are carried out on PASCAL VOC 2007 and 2012. Our OPG shows effective improvement on the performance of our baseline, setting a  performance comparable to the state-of-the-arts.

\section{Related Work} 

\subsection{Weakly Supervised Object Detection}

Recently, weakly supervised object detection (WSOD) has been extensively studied. Most existing methods \cite{oquab2015object,bilen2016weakly,tang2017multiple,tang2018weakly,zhang2018zigzag,wan2019c,kosugi2019object,shen2019cyclic,gao2019c,wei2018ts2c,kantorov2016contextlocnet} follows into a two-step training process. They mainly focus on solving the problems at the first step.

In this first step, they formulate the WSOD task as a Multiple Instance Learning (MIL) problem. WSDDN \cite{bilen2016weakly} proposes a Convolutional Neural Network (CNN) based frameworks which adopts a two stream construction and uses a spatial regulariser to further improve the performance. Based on the methods \cite{bilen2016weakly,oquab2015object}, ContextLocNet \cite{kantorov2016contextlocnet} propose a context-aware model which combines the context region to improve the detection accuracy. OICR \cite{tang2017multiple} develops an online refinement algorithm to further improve the performance by using spatial information. The approach proposed in \cite{kosugi2019object} labels objects in instance-level using spatial constrain. 
WSOD2 \cite{zeng2019wsod2} develops a box regressor based on the fusion of bottom-up and top-down features.
The performances of MIL based methods are limited by the non-convexity of their loss functions. To mitigate this issue, C-MIL \cite{wan2019c} proposes a continuation algorithm to improve the detection accuracy. Weakly supervised segmentation methods are widely applied in some recent works. The methods in \cite{shen2019cyclic,gao2019c,wei2018ts2c} jointly train detection and segmentation frameworks under weak supervision to improve their performances. 

At the second step, people select the top scored proposals by the trained network at the first step as pseudo ground truths (PGTs). Another network is trained with these PGTs under full supervision for better performance.

Recently, there are two works \cite{tang2018weakly,singh2019you} whose aims are similar to our OPG algorithm's. The method in \cite{tang2018weakly} proposes a region proposal network (RPN) for WSOD. Its training process, which consists of three stages, is time-consuming and complex. The RPN proposed in it requires alternative training stage by stage. Without end-to-end training, its performance is not satisfactory.  With an external video dataset, W-RPN \cite{singh2019you} proposes another RPN to improves the quality of proposals in weakly supervised detection methods. It utilizes the motion cues existing in videos to generate high precision proposals. 

Different from them, our OPG algorithm uses the prediction of the training network to generate the active training proposal set in an online manner. Our OPG algorithm does not require alternative training procedure or any external dataset. Since the predictions of the training network can be easily obtained during the training process, our OPG algorithm does not introduce much computation burden on training. Our OPG algorithm can be easily applied to any weakly supervised object detection methods.

\subsection{Hard Example Mine Method}

Our proposed method can be also regarded as an approach to mine the hard examples during the training process. Compared with weakly supervised object detection methods, there are many approaches \cite{lin2018focal,shrivastava2016training,pang2019libra} proposed to mine hard examples under full supervision.

OHEM \cite{shrivastava2016training} sorts proposal samples by loss based and selects a certain number of samples as the hard examples for training. In RetinaNet \cite{lin2018focal}, it firstly filters out some samples according to their overlaps with the box-level annotations and focuses on training the retained samples with large loss. In the method \cite{pang2019libra}, it tries to solve this problem by computing the overlaps between these samples and the box-level ground truths. Among these methods, the box-level annotations are necessary components. Since box-level annotations are not available in weakly supervised object detection, it is challenging to apply these methods to the weakly supervised object detection task. To overcome this issue, we propose an online active proposal set generation algorithm without the needing of box-level annotations.

\section{Ratio Influence on Detection}\label{sec3}

The proposal set is dominated by negative samples. Without box-level annotations, existing weakly supervised object detection methods use all proposals for training. This may degrade their performances. This issue is neglected by these methods. To show its importance, we investigate the impact of the ratio between positive and negative samples on detection accuracy in this section.

\begin{figure*}[htbp]
	\centering
	{\includegraphics[width=\textwidth]{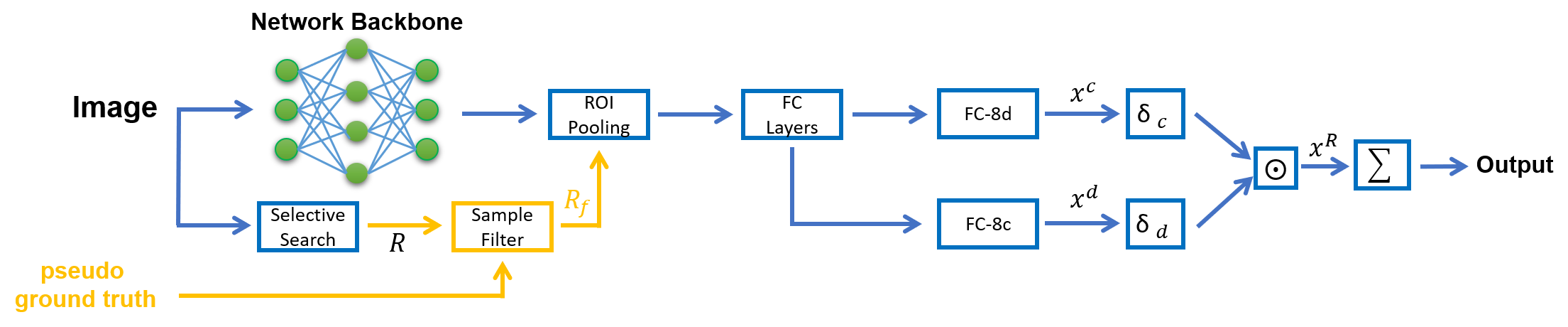}
	}
	\caption{Illustration of the basic detection framework under weak supervision.}
	\label{fig:basic_w}
\end{figure*} 

This series of experiments are conducted based on a framework \cite{bilen2016weakly} as shown in Fig. \ref{fig:basic_w}. Following WSDDN \cite{bilen2016weakly}, the proposal features are connected with two layers: FC-8d and FC-8c. Then, outputs, $\textbf{x}^{c}$ and $\textbf{x}^{d}$, are forwarded to two softmax functions, respectively: $\delta_{c}(\textbf{x}^{c})(i,j) = \frac{e^{x^{c}_{i,j}}}{\sum_{m=1}^{C}e^{x^{c}_{m,j}}}$, and  
$\delta_{d}(\textbf{x}^{d})(i,j) = \frac{e^{x^{d}_{i,j}}}{\sum_{m=1}^{|R_{r}|}e^{x^{d}_{i,m}}}$,
where $|R_{r}|$ denotes the number of the actually forwarded proposals and $C$ is the number of classes.
The final score for each proposal is computed by an element-wise production: $\textbf{x}^{R} = \delta_{c}(\textbf{x}^{c}) \odot \delta_{d}(\textbf{x}^{d})$. Finally, the proposal scores at $c$ th class are summed up: $p_{c} = \sum_{k=1}^{|R_{r}|}x^{R}_{ck}$. We train this network using a cross entropy loss $L_{base}$, as listed in Eq. \ref{loss_1}.
\begin{equation}
L_{base} = -\frac{1}{|C|}\sum_{c=1}^{C}[y_{c}\ln p_{c}+(1-y_{c}\ln(1-p_{c}))],\label{loss_1}
\end{equation}where $y_{c}$ is image level annotation for class $c$ and $C$ is the number of classes.

To control the ratio between positive and negative samples in the proposal set, we firstly train the network with only blue parts as shown in Fig. \ref{fig:basic_w} under weak supervision, obtaining the PGTs. After that, without sharing parameters with the trained model above, we train new networks including yellow parts and change the ratio according to the generated PGTs. To ensure enough proposals for training, we use a fixed number of proposals for each experiment and keep the ratio around a pre-defined value.

\begin{figure}[htbp]
	\centering
	{\includegraphics[width=.6\textwidth]{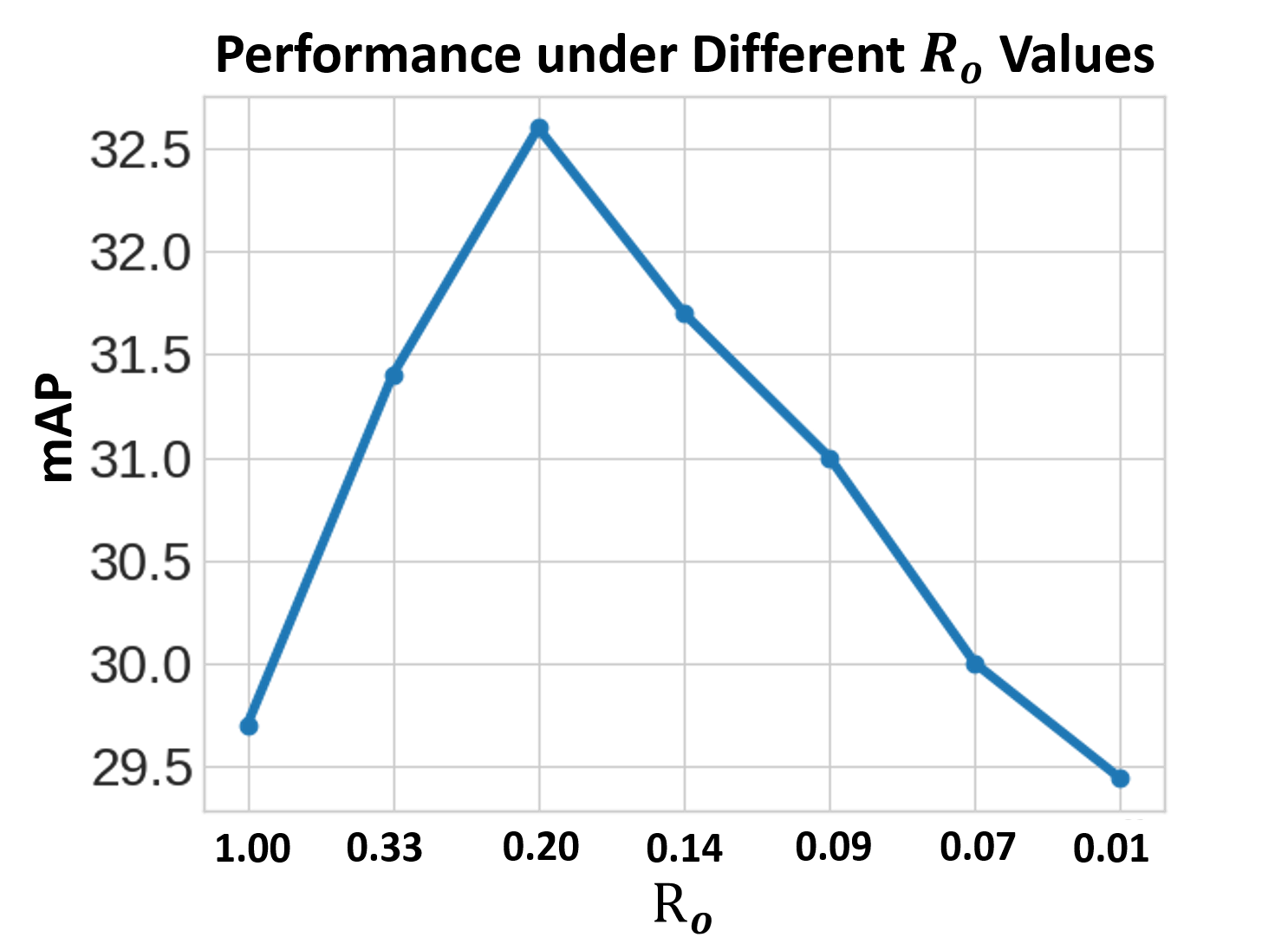}
	}
	\caption{The performances of the trained models under different $R_{o}$. This shows that the unreasonable ratio in existing weakly supervised methods limits their performances.}
	\label{fig:c_exp}
\end{figure}

Seven experiments under different ratios are conducted, where $R_o = |P_{f}|/|P_{b}|$ indicates the ratio between positive and negative proposals. The performances of the trained models under different $R_o$ are plot in Fig. \ref{fig:c_exp}.   

From Fig. \ref{fig:c_exp}, it can be seen that the performance of the trained model is improved when $R_{o}$ decreases from 1.00 to 0.20. When $R_{o}$ continues to decrease, the performances is gradually degraded. It is because the proposal set under ratio $R_{o}$ between 0.2 and 1.0 is not dominated by the negative proposals. For smaller $R_{o}$, the proposal set is dominated by negative samples, degenerating the detection accuracy. For the experiment with $R_{o} = 0.01$, all proposals are involved in training, the performance is dramatically decreased. This experiment reveals that the proposal ratio significantly affects the performances of weakly supervised object detection methods.

In these experiments above, we use the PGTs from the training network parts in blue to divide proposals into positive and negative samples. Instead of using these PGTs, we can also manually make box-level annotations and use these box-level annotations to control the proposal ratio in these experiments above. \textit{Our aim is to solve this imbalanced problem without box-level annotations. To avoid the influence introduced by the manual box-level annotations, we do not use this manual box-level annotation in this section.}

\section{Online Active Proposal Set Generation}

\begin{figure*}[htbp]
	\centering
	{\includegraphics[width=\textwidth]{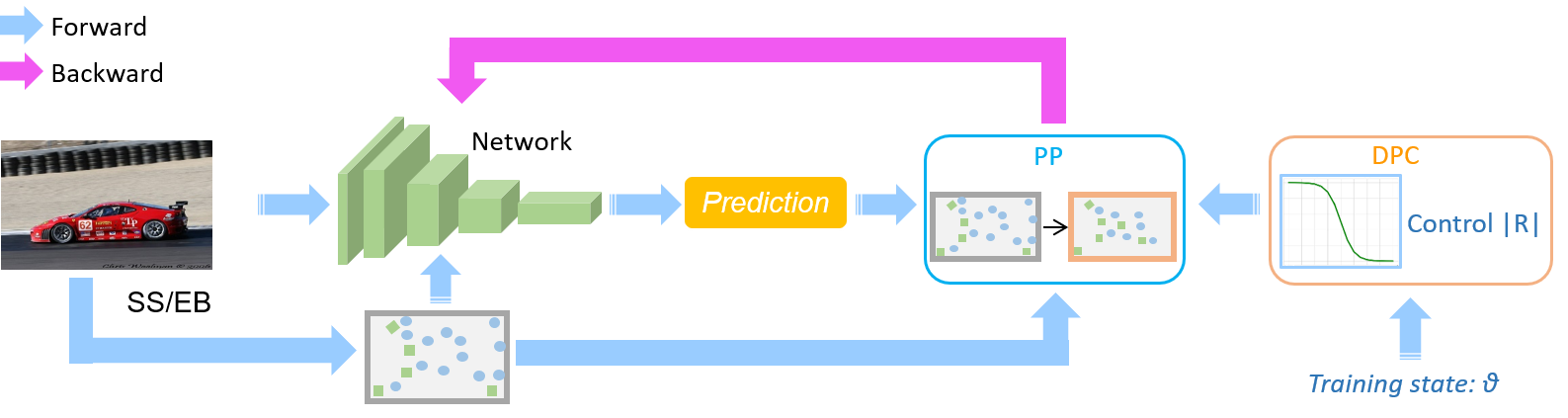}
	}
	\caption{The illustration of our \textbf{O}nline Active \textbf{P}roposal Set \textbf{G}eneration (OPG) algorithm. Our OPG algorithm consists of two components: dynamic proposal constraint (DPC) and proposal partition (PP). In the training process, forward propagation and backward propagation are indicted by blue and purple anchor, respectively. At the forward stage, we pass all proposals to the training network. DPC determines the proposal sampling strategy according to a training state $\theta$, which increases from 0 to 1, during training. PP employs the sampling strategy from DPC and generates an active proposal set based on the training network predictions. At backward stage, we only optimize the training network on the active proposal set.}
	\label{fig:overview}
\end{figure*}

According to Section \ref{sec3}, we find that the proposal ratio hinders the detection performance for weakly supervised detection methods. Under weak supervision, the training network especially for first several epochs is easily to get stuck into a local minima, producing many false PGTs. A simple proposal sampling method is not applicable. To address this issue, we propose an \textbf{O}nline Active \textbf{P}roposal Set \textbf{G}eneration (OPG) algorithm. Our OPG consists of two components: dynamic proposal constraint (DPC) and proposal partition (PP). The overview for our OPG algorithm is illustrated in Fig. \ref{fig:overview}.

As illustrated in Fig. \ref{fig:overview}, we divide the training process into two stage: forward propagation stage and backward propagation stage. We use blue and purple anchor to indict the forward and backward stage, respectively. At the forward stage, we pass an image to the training network and selective search \cite{uijlings2013selective} or edge box \cite{zitnick2014edge} to predict scores for each proposals. Then, our DPC determines the proposal sample strategy based on a training state $\theta$, which increase from 0 to 1 during the training process. Our proposed PP generate the active proposal set according to the training network prediction and the proposal strategy from DPC. At the backward stage, we only optimize the training network within the active proposal set. 

\subsection{Dynamic Proposal Constraint} \label{subsec:dpc}

During the training process, the training network performance varies. This affects the quality of the online PGTs from this training network.
Proposal sampling based on low quality PGTs may remove many positive proposals and destroy the training process. To solve this problem, we expect that the training network frees from proposal sampling based on low quality PGTs, while benefiting from proposal sampling based on high quality PGTs. To realize this aim, a Dynamic Proposal Constraint (DPC) is proposed.

DPC is able to determine different proposal sampling strategies according to the current training state. We define a training state $\theta$. During the training process, $\theta$ is 0 at the beginning and increases to 1 in the end. In DPC, different proposal sampling strategies should be well adapted to the current performance of the training network. In view of the training network performance, we divide the training process into three stages:

\begin{itemize}
	
	\item warm up stage: The training network cannot identify the positive and negative proposals accurately. PGTs produced by this network involve many false true results and cannot be used to filter out redundant negative proposals.
	
	\item transition stage: The training network performance is gradually improved. Some easy samples can be detected. Proposal sampling based on PGTs from this network may be beneficial to the training process
	
	\item stable stage: The training network is able to identify most samples. Propolsa sampling based on the PGTs generated from this network can be used for performance improvement.
	
\end{itemize}

According to the characters among the three training stages, we determine three different proposal sampling strategies. At the warm up stage, we use all proposals in training. Since no proposals are removed, the training network is improved stably. Then, at the transition stage, we gradually removes the number of negative proposals and fuurther improve the training network performance. Finally, at the stable stage, we remove most negative proposals. This prevent the training network from biasing to negative proposals and enable the training network to achieve a better performance. 

\begin{figure}[htbp]
	\centering
	{\includegraphics[width=.6\textwidth]{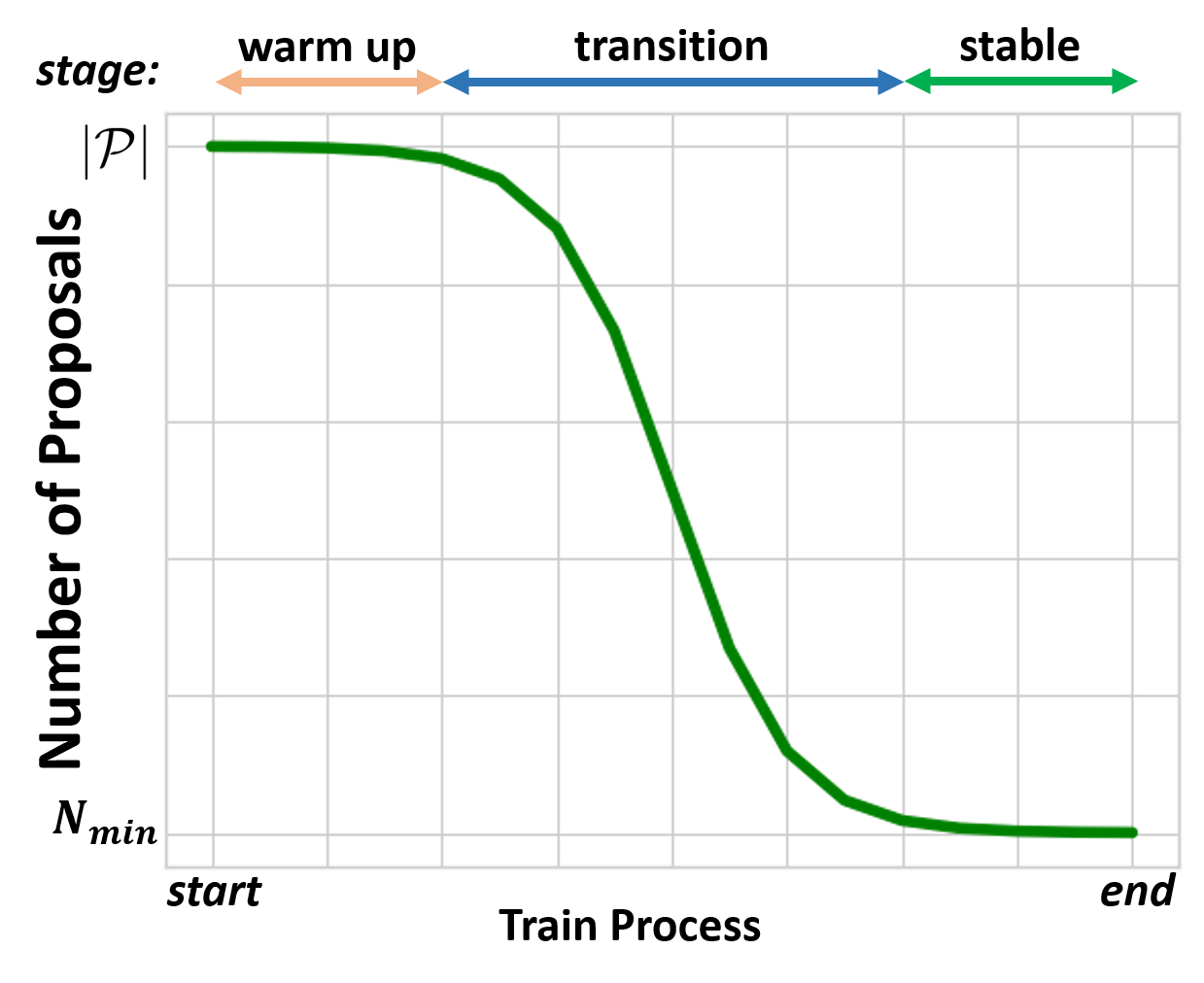}
	}
	\caption{Proposal numbers in our active training proposal set for different training stages.}
	\label{fig:loss}
\end{figure}

We formulate these three proposal sampling strategies as a single equation, which is written as follows:  
\begin{equation}
\gamma(\theta) = \frac{1}{1+e^{\alpha * (\omega*\theta - \beta)}}\label{eq:gamma},
\end{equation}where $\theta$ is the training state, $\alpha$, $\beta$ and $\omega$ are hyper-parameters. The proposal number $N_{v}(t)$ in our active training proposal set is defined as:
\begin{equation}
N_{v}(\theta) = \max(\gamma(\theta)*|\mathcal{P}|, N_{min}),\label{eq:nv}
\end{equation}where $|\mathcal{P}|$ denotes the total number of original proposals, $N_{min}$ is a constant. 

After employing our DPC, the $N_{v}(\theta)$ during training process is plot in Fig. \ref{fig:loss}. From the Fig. \ref{fig:loss}, we can find that nearly all proposals are involved at the warp up stage. The number of proposal at the transition stage gradually decreases. Finally, the number of proposals at the stable proposal decreases to the minimum value, removing the overwhelming negative proposals.

It is not the only way to implement our proposed algorithm. The core behind our proposed algorithm is to understand the network detection accuracy variation during the training process and know how to remove those redundant negative proposals while retaining most positive proposals. \textit{Our experiments reveal that our proposed algorithm shows an excellent improvement when the warm up stage and the stable stage occupies around 40\% and 50\%, respectively in the training process.} It can also be realized by other implementation methods, if these training stages are assigned under the condition discussed above. 

\subsection{Proposal Partition} 

Given the proposal sampling strategy from our DPC, our proposal partition (PP) is used to score each proposal, part proposals into different sets and generate the active proposal set. Our PP is provided in Alg. \ref{alg:pp}.

\begin{algorithm}[htbp]
	\SetAlgoLined
	\KwInput{Image \textbf{I}, class label \textit{y}, proposals \textbf{R}, threshold $t_f$,$t_{b1}$, $t_{b2}$, $N_{v}$ from DPC }
	\KwOutput{The active proposal set $P_a$}
	Feed \textbf{I} into the training network; get ROI scores \textbf{S}\;
	\For{ground-truch class $c$}{
		$r_{g} \leftarrow \max(\textbf{S}(c,:))$ \tcp*{select a PGT}\
		$\textbf{G}$ $\leftarrow$ $r_g$
		\tcp*{save this PGT to set \textbf{G}}
	}
	\For{proposal $r$}{
		$s_p$ $\leftarrow$ $\arg\max_{c}$IOU($r$, \textbf{G})
		\tcp*{compute the proposal scores}\
		$\textbf{S}^{'} \leftarrow s_p$ \tcp*{save the score to set $\textbf{S}^{'}$}
	}
	Separate set \textbf{R} into three sets, $P_f$, $P_b$ and $P_r$ according to Eq \ref{eq:iou}\;
	Compute $N_p$ and $N_b$ according to Eq \ref{eq:np} and Eq \ref{eq:nb} \;
	$P_f^s \leftarrow \rm{SORT}(P_f), P_b^s \leftarrow \rm{SORT}(P_b)$\tcp*{sort set}\
	$P_f^{'}$ $\leftarrow$ top $N_{p}$ of $P_f^s$, $P_b^{'} \leftarrow$ top $N_{b}$ of $P_b^s$\;
	$P_a \leftarrow P_f^{'}\cup P_b^{'}$\;
	\textbf{if} $|P_a|$ \textless $N_v$:
	$P_u^r$ $\leftarrow$ RANDOM($P_u$, $N_v$-$|P_a|$), APPEND ($P_a$,$P_u^r$)
	\caption{Proposal Partition}
	\label{alg:pp}
\end{algorithm}

Suppose the image-level annotation for an image is $\textbf{y} = [y_{1}, y_{2}, ..., y_{C}]^{T} \in \mathbb{R}^{C\times1}$, where $y_c$ is 1 or 0 denotes whether this image includes the class $c$. For each training iteration, we firstly choose the prediction with highest score at class $c$ as a pseudo ground truth $r_g$, if its corresponding annotation $y_c$ is 1. Then, we save all $r_g$ in a set \textbf{G} and compute the propose score $s_{p}$ for each proposal $r$ according to Eq. \ref{eq:rc}.

\begin{equation}\label{eq:rc}
s_{p} = {\rm{arg}}\max_{c}{\rm{IOU}}(r, \textbf{G})
\end{equation}After that, we part the original proposal set into three set: positive set $P_f$, negative set $P_b$ and risk set $P_r$ as follows:	
\begin{equation}\label{eq:iou}
l_{r} =
\begin{cases}
1 & s_p \geq t_f\\
0 & t_{b1} > s_p \geq t_{b2}\\
-1 & \rm{others}\\
\end{cases}  
\end{equation}where $l_{r}$ denotes the proposal $r$ label, $\theta_f$, $\theta_{b1}$ and $\theta_{b2}$ are constants.
$l_r = -1$ indicates the proposal is separated into a risk set. The risk set is used to express some hard samples for the training network to classification. 

The number of positive proposals, $N_{p}$ and negative proposals, $N_{b}$ are calculated according to Eq. \ref{eq:np} and Eq. \ref{eq:nb} , respectively. Then, we sort proposals in the positive and the negative proposal set, $P_f$ and $P_b$, respectively. The top $N_p$ proposals in $P_f$ and the top $N_b$ proposals in $P_b$ are selected and formed into our active proposal set $P_a$. Sometimes the proposal number in $P_f$ may be smaller than $N_p$, resulting in $|P_a| < N_v$. In this case, we randomly select $N_v - |P_a|$ proposals from the risk set $P_r$ and append them to $P_a$.

\begin{equation}
N_{p} = r_p * N_{v}
\label{eq:np}
\end{equation}

\begin{equation}
N_{b} = (1-r_p) * N_{v}
\label{eq:nb}
\end{equation}

\textbf{Reason to define a risk set.} Some prople may be confused by the risk set definition. To present our PP clearly, we explain it in details. Negative proposals can be further divided into two classes: region with object parts and region without object parts (pure background proposals). During our experiments, we find that networks trained with many pure background proposals show inferior performances than other networks trained without these proposals. Maybe it is because networks trained with these pure background proposals are more easily prone to negative parts. To alleviate this issue, we define the risk set and try to train a network without these proposals. 

\subsection{Detection Framework}\label{sec4-}

\begin{figure*}[htbp]
	\centering
	{\includegraphics[width=\textwidth]{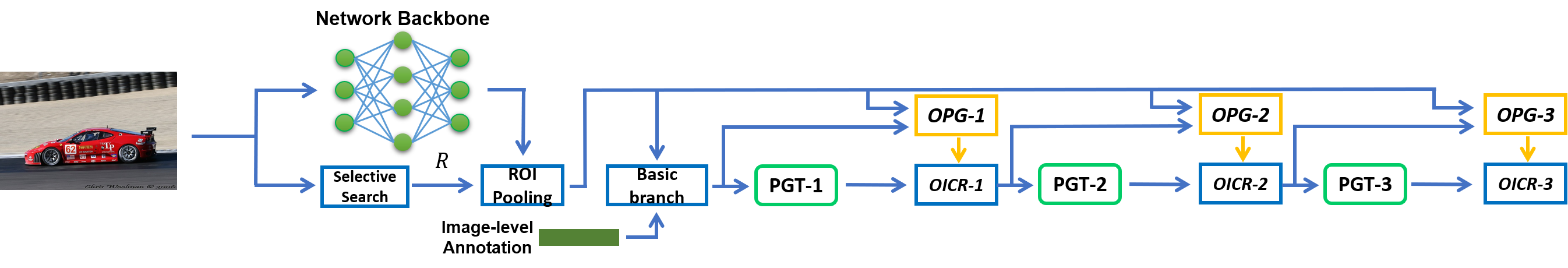}
	}
	\caption{Illustration of our weakly supervised object detection framework.}
	\label{fig:detect}
\end{figure*}

We adopt the OICR \cite{tang2017multiple} as our detection framework which is illustrated in Fig. \ref{fig:detect}. Our detection
framework consists of four branches: one basic branch and three OICR branches. The basic branch is the same as FC layers in Fig. \ref{fig:basic_w}, which is has been presented in Section \ref{sec3}. For the OICR branche, it regards the predictions with highest score as pseudo ground truths and divides proposals into positive and negative samples, which is similar the step 1 in our proposal ranking (PR). The only different point is that the OICR branch merges the unknown proposals in Eq. \ref{eq:iou} with the negative samples together. Given the predicted scores of the pseudo ground truths $w_{ik}$ for the $i$th proposal at the $k$th OICR branch, the loss function $L_{r}^{k}$ at the $k$th OICR branch is defined as:

\begin{equation}
L_{r}^{k} = -\frac{1}{|R|}\sum_{i=1}^{R}\sum_{c=1}^{C+1}w_{ik}y_{ik}^{c}\ln x_{ik}^{c},
\end{equation}where $y_{ik}^{c}$ is the class $c$ label for the $i$th proposal at the $k$ th OICR branch.

In the Figure \ref{fig:detect}, we firstly generate the pseudo ground truths, PGT-1 according to the prediction of the basic branch and pass PGT-1 to OICR-1. Simultaneously, our OPG-1 generates the active training proposal set for the current training stage and pass this proposal set the OICR-1. We repeat these operations iteratively for the following two OICR branches. Finally, the prediction from three OICR branches are averaged to obtain the final results. The loss function for our detection framework is defined as:

\begin{equation}
L = L_{base} + \sum_{m=1}^{M}L_{r}^{m},
\end{equation}where M is 3.

\section{Experiments}

\subsection{Experiment Settings}
\textbf{Datasets and evaluation metrics} Following the protocol of previous methods \cite{tang2017multiple,kantorov2016contextlocnet}, our methods are evaluated on the benchmark PASCAL VOC 2007 and 2012 datasets \cite{everingham2010pascal}. We choose the trainval set as our training dataset and use the testing set as our testing dataset. Two metrics, mean average precision (mAP) and correct localization (CorLoc), are used evaluation. Following the protocol of PASCAL VOC \cite{everingham2010pascal}, mAP is used to measure our method performance on the testing set. CorLoc is to evaluate our method performance on the training set. The threshold value for IOU in both evaluation metrics is 0.5.

\textbf{Implementation Details} The network backbone is VGG16 \cite{simonyan2014very} which is pre-trained on ImageNet \cite{deng2009imagenet} dataset. We replace the last max-pooling layer, $pool5$ with a ROI-pooling layer \cite{girshick2015fast}. To improve the resolution of feature map, we remove the penultimate pooling layer, $pool4$ and increase the dilation rate on the conv5 layers from 1 to 2. For training, SGD optimization method is performed and the mini-batch size is 2. The learning rate for the first 40K iterations is set to 0.001 and then decreases to 0.0001 for the following 55K iterations. The momentum is set to 0.9 and the weight decay is 0.0005.  

According to previous methods \cite{tang2017multiple,wan2019c}, multi-scale training and testing are adopt in our experiments. we use five images scales and resize the shorter side of an image to one of these sizes: 480, 576, 688, 864, and 1200. The horizontal flip is also used for data argumentation. Selective search \cite{uijlings2013selective} is applied to generate the original proposals. For the hyper-parameters in OICR, we follow the original setting in OICR \cite{tang2017multiple}. For the hyper-parameters in OPG, $r_p$ is 0.25, $t_f$ and $t_{b1}$ are 0.5 and $t_{b2}$ is 0.1.  

\subsection{Analysis of OPG Algorithm}

In this part, we discuss the effectiveness of the hyper-parameters in Eq \ref{eq:gamma}  on the experiments. In this part, since $\omega$ shows similar influence to $\alpha$, we fix $\omega$ at 1.36 and only change the value of $\alpha$ and $\beta$. As discussed in Section \ref{sec4-}, we directly apply our OPG algorithm to OICR \cite{tang2017multiple} and use OICR as our baseline. We evaluate mAP on VOC 2007 test set and CorLoc on VOC 2007 trainval set. We only apply multi-scale in training and resize the shorter size of an image to a single size, 1200 in testing.

\begin{figure}[htbp]
	\centering
	\begin{subfigure}[]{
			\label{fig:plot1} 
			\includegraphics[width=.45\textwidth]{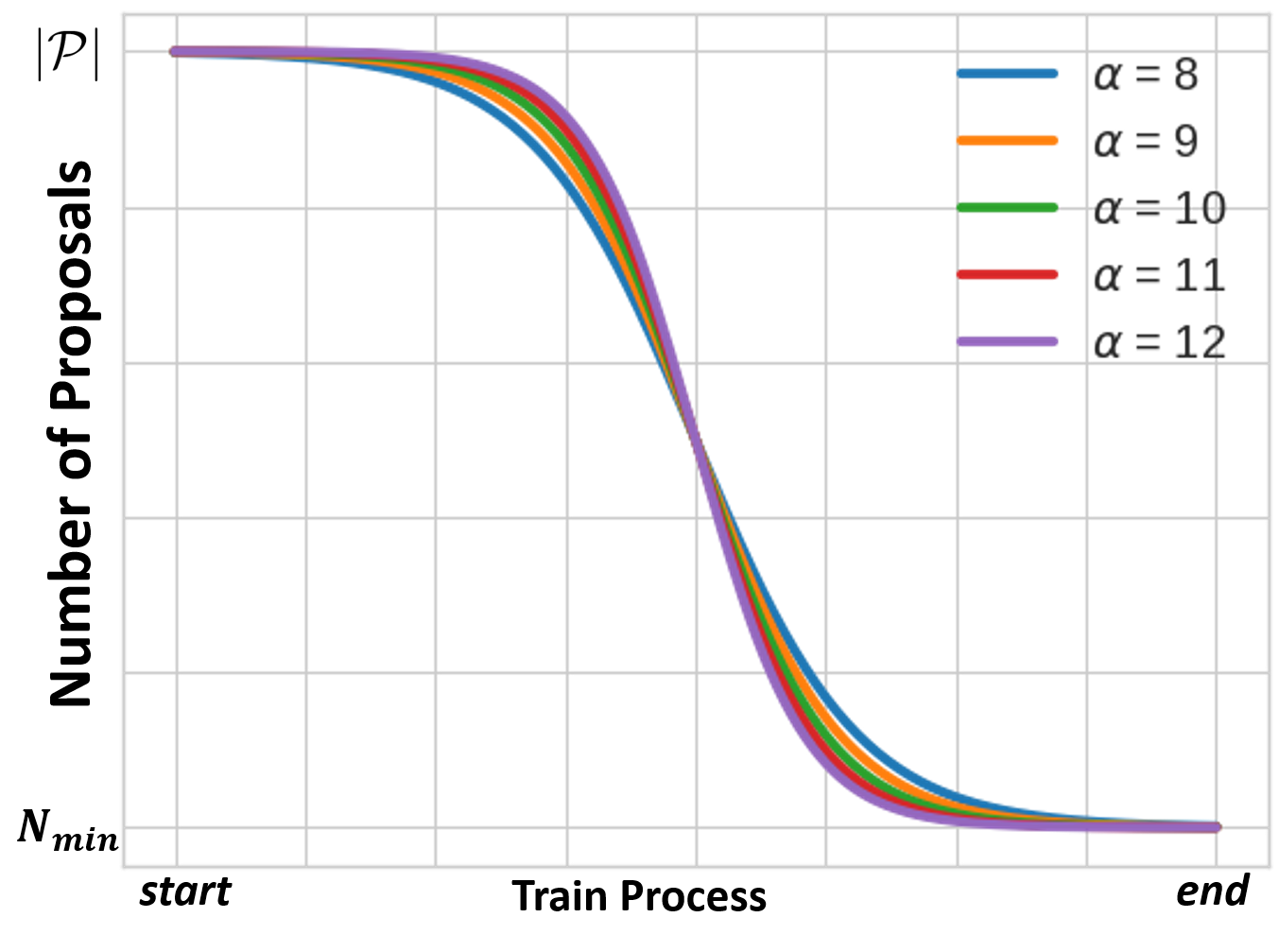}} 
	\end{subfigure}
	\begin{subfigure}[]{
			\label{fig:plot2}
			\includegraphics[width=.45\textwidth]{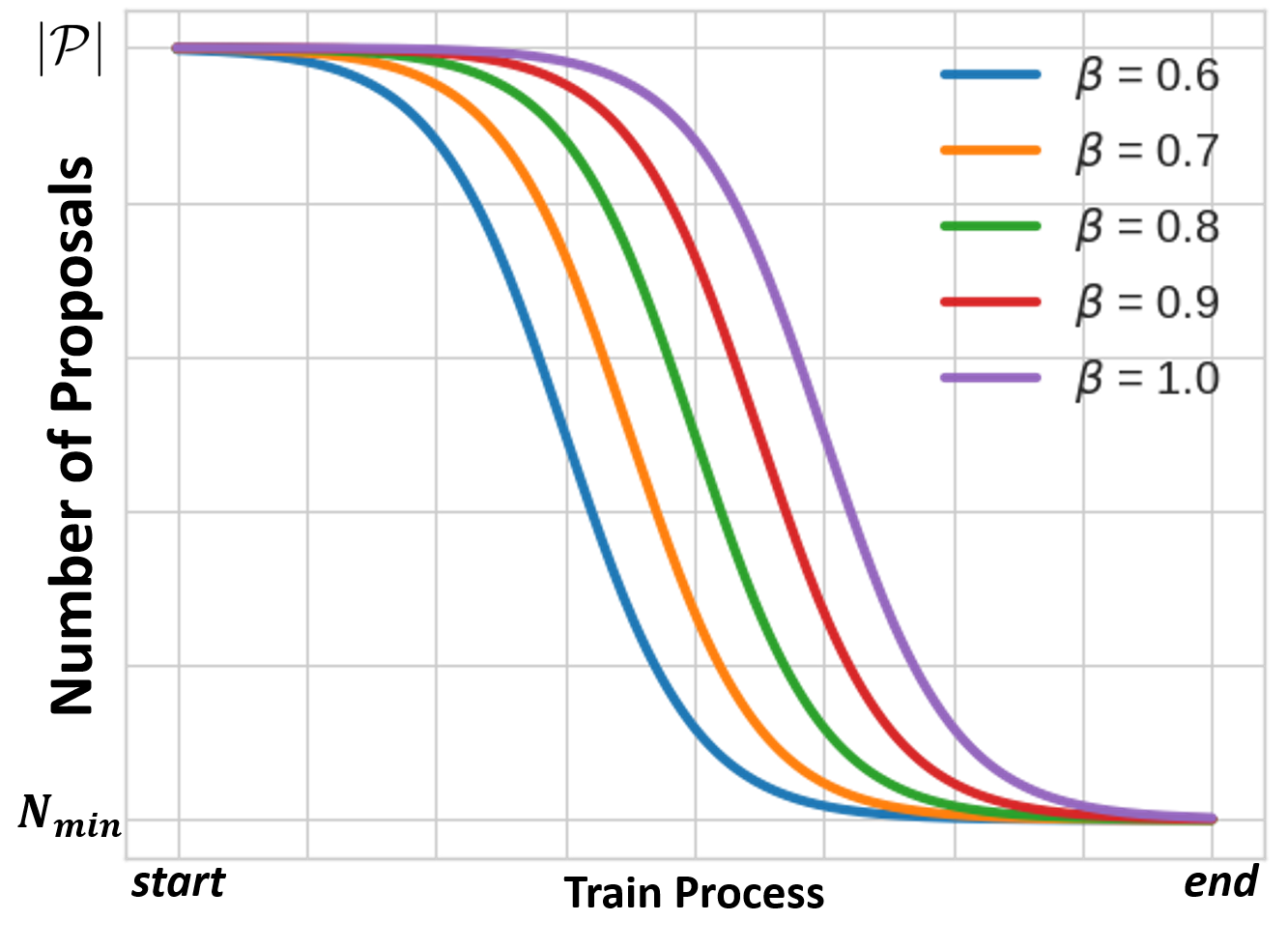}} 
	\end{subfigure} 
	\caption{Training proposal number with our OPG under different $\beta$  and $\alpha$ values.}
\end{figure} 

The $N_{v}(\theta)$ for different $\alpha$ values are plot in Fig. \ref{fig:plot1}. From the Fig. \ref{fig:plot1}, it can be seen that $\alpha$ controls the occupation of the transition stage in training. Results for various $\alpha$ are listed in Table \ref{table:exp1}, where $\beta$ is fixed at 0.8. It shows that nearly all models trained with OPG perform better than our baseline. When $\alpha$ is 10.0, the model trained with OPG yields the best performance. Some experiments for various $\beta$ are also conducted and the corresponding $N_{v}(t)$ is plot in Fig. \ref{fig:plot2}. In Fig. \ref{fig:plot2}, it shows that the $\beta$ changes the ratio between the epochs in the warm up stage and that in the stable stage. The results for different $\beta$ are shown in Table \ref{table:exp2}, where $\alpha$ is fixed to 10.0. From the Table \ref{table:exp2}, it demonstrates that our OPG effectively improves the performance of the baseline. When $\beta$ is equal to 0.8, the model trained with our OPG shows the best performance.

\begin{table}[htbp]
	\caption{Experiments for various $\alpha$ on VOC 2007, where $\beta$ is fixed at 0.8. It shows that nearly all model trained with our OPG show better performances than our baseline.}
	\begin{center}
			\begin{tabular}{c|c|c c}
				Method &	$\alpha$ & mAP & CorLoc  \\
				\hline
				Baseline (OICR)    &      -      & 36.7&  61.4   \\
				Baseline+OPG (ours)&	 8.0     & 45.9&  63.7   \\
				Baseline+OPG (ours)&	 9.0     & 45.8&  63.4   \\
				Baseline+OPG (ours)&	 10.0    & \textbf{46.0}&  \textbf{64.1}   \\
				Baseline+OPG (ours)&	 11.0    & 44.6&  61.2   \\
				Baseline+OPG (ours)&	 12.0    & 45.5&  63.1   \\ 
		\end{tabular}
	\end{center}
	\label{table:exp1}
\end{table}

In the analysis experiments above, our OPG effectively improves
the baseline. The performance of our OPG is robust to the variation of $\alpha$ and $\beta$. In the following experiments, $\alpha$ and $\beta$ are 10.0 and 0.8, respectively.

\begin{table}[htbp]
	\caption{Experiments for various $\beta$ on VOC 2007, where $\alpha$ is fixed at 10.0. It shows that all model trained with our OPG show better performances than our baseline.}
	\begin{center}
			\begin{tabular}{c|c|c c}
				Method &	$\beta$ & mAP & CorLoc  \\
				\hline
				Baseline (OICR)&      -    &36.7 &  61.4 \\
				Baseline+OPG (ours)&	0.6     & 43.9&  63.9   \\
				Baseline+OPG (ours)&	0.7     & 45.0&  63.0   \\
				Baseline+OPG (ours)&	0.8    & \textbf{46.0}&  \textbf{64.1}   \\
				Baseline+OPG (ours)&	0.9    & 45.4&  63.7   \\
				Baseline+OPG (ours)&	1.0    & 42.6&  62.5   \\ 
		\end{tabular}
	\end{center}
	\label{table:exp2}
\end{table}

We would like to point out that this subsection is to verify our proposed algorithm is robust to the variation of $\alpha$ and $\beta$. Although these experiments are conducted with the Eq. \ref{eq:gamma}, it is not the only implementation for our proposed algorithm. In this paper, we provide an example implementation for our proposed method. As presented in \ref{subsec:dpc}, our proposed method shows significant improvement on the condition that the warm up stage and the stable stage occupies around 40\% and 50\%, respectively in the training process. It is easily realized by other implementation methods, if these training stages are assigned under the condition discussed above.

\subsection{Effectiveness of OPG algorithm}

In this part, we conduct experiments to verify the effectiveness of our OPG algorithm. We firstly re-implement OICR by ourselves, which performs a little better than the original OICR. We compare the performance of OICR and our detection method OICR + OPG with mAP and CorLoc in Table \ref{table:ablat1} and Table \ref{table:ablat2}, respectively. 

\begin{table*}[htbp]
	\caption{Compare our OPG method with the baseline in \textit{mAP} on PASCAL VOC 2007 \textit{test} set. Our OPG algorithm improves the baseline performance in nearly all classes.}
	\begin{center}
		\scalebox{0.68}{
			\begin{tabular}{|c|c c c c c c c c c c |c|}
				\hline
				Method & aero &bike& bird & boat & bottl & bus & car & cat & chair & cow & \\ 
				\hline
				Baseline (OICR)& 55.7 &63.1& 37.4 & 19.4 & 17.2 & 64.7 & 65.9& 43.9& \textbf{27.7} & 48.7 &\\
				Baseline+OPG (ours)    &\textbf{63.0} &\textbf{65.3 }& \textbf{49.2}& \textbf{31.7} & \textbf{25.3} & \textbf{70.9} & \textbf{70.9}& \textbf{58.1} &27.4& \textbf{58.6} & \\
				\hline
				Method & table & dog & horse & mbike & per&plant&sheep&sofa& train& tv& mean \\
				\hline
				Baseline (OICR)& 35.5 & 35.8 & 40.6  & 61.4  &8.9&24.2& 40.6& 46.9& \textbf{62.3}&64.2&43.2\\
				Baseline+OPG (ours)	&  \textbf{44.7}&  \textbf{47.0}& \textbf{47.2}  & \textbf{69.8}  &\textbf{13.1}&\textbf{26.1}& \textbf{49.9}& \textbf{51.8}& 61.7&\textbf{68.2}& \textbf{50.0}\\	
				\hline		
		\end{tabular}}
	\end{center}
	\label{table:ablat1}
\end{table*}

\begin{table*}[htbp]
	\caption{Compare our OPG method with the baseline in \textit{CorLoc} on PASCAL VOC 2007 \textit{trainval} set. It shows that our OPG improves the performance of the baseline.}
	\begin{center}
		\scalebox{0.68}{
			\begin{tabular}{|c|c c c c c c c c c c|c|}
				\hline
				Method      & aero &bike& bird & boat & bottl & bus & car & cat & chair & cow & \\
				\hline
				Baseline (OICR)&82.1&\textbf{83.9}&\textbf{64.9}&49.5&\textbf{43.5}&\textbf{81.2}&85.9&56.4&\textbf{44.1}&75.3& \\
				Baseline+OPG (ours)    &\textbf{83.3}&77.7&64.0&\textbf{50.0}&43.1&80.7&\textbf{87.5}&\textbf{61.6}&40.7&\textbf{79.5}& \\
				\hline
				Method & table & dog & horse & mbike & per&plant&sheep&sofa& train& tv& mean\\
				\hline
				Baseline (OICR) &\textbf{43.7}&53.3&61.6&88.4&14.7&57.9&\textbf{78.4}&57.5&\textbf{77.6}&79.6&64.0 \\
				Baseline+OPG (ours)
				&43.4&\textbf{60.0}&\textbf{71.4}&\textbf{90.8}&\textbf{15.9}&\textbf{58.6}&76.3&\textbf{62.6}&74.9&\textbf{83.2}&\textbf{65.3} \\
				\hline
		\end{tabular}}
	\end{center}
	\label{table:ablat2}
\end{table*}

In Table \ref{table:ablat1}, it shows that our method (OICR+OPG) performs significantly better than the baseline OICR (50.0 v.s 43.2) in mAP. Trained with our OPG, OICR+OPG surpasses OICR in \textit{nearly all classes}. In the Table \ref{table:ablat2}, our OPG algorithm improves the performance of the baseline (65.3 v.s. 64.0) in CorLoc. This demonstrates that out OPG mitigates the issue caused by the imbalance of positive and negative samples in the training proposal set. 

\begin{table}[htbp]
	\caption{Comparison with OICR+FRCNN. OICR+FRCNN denotes a Fast RCNN trained with PGTs from OICR network at the second step training. It shows that our proposed OPG algorithm still performs better than the OICR+FRCNN within only one step.}
	\begin{center}
			\begin{tabular}{c|c c|c c}
				\hline
				Dataset &\multicolumn{2}{c}{VOC 2007}&\multicolumn{2}{|c}{VOC 2012}\\
				\hline
				Method & mAP   & CorLoc  & mAP  &CorLoc\\
				\hline
				OICR+FRCNN\cite{tang2017multiple}    & 47.0 &  64.8 & 42.5& 65.6\\
				\hline
				OPG (ours)  & \textbf{50.0} &    \textbf{65.3} & \textbf{46.2} &  \textbf{65.8}\\
				\hline				
		\end{tabular}
	\end{center}
	\label{table:cmp1}
\end{table}

Our proposed OPG algorithm aims to combine the two-step training into one-step training, providing an effective and efficient training methods. To show the efficiency of our OPG algorithm, we compare our OPG method with OICR+FRCNN at table \ref{table:cmp1}. From the table \ref{table:cmp1}, although OICR+FRCNN uses two-step training, its performance is still inferior to our OPG. This demonstrates that our proposed OPG is able to effectively combine the two-step training into one-step training, while showing a better performance. 

\subsection{Comparison with the state-of-the-art}

In this subsection, we compare our method with other state-of-the-arts in Table \ref{table:other2}. As listed in table \ref{table:other2}, our method shows a comparable state-of-the-art results. Since our proposed OPG algorithm is directly applied to OICR method, the performance od OICR+OPG (ours) is limited by the OICR method. For improvement, our proposed OPG significantly improves our baseline by around 7 mAP. This improvement is larger than most methods. For example, C-MIDN \cite{gao2019c} shows an excellent performance in VOC 2007 (53.6 mAP and 68.7 CorLoc). But it benefits a strong baseline which obtains 49.0 mAP and 46.8 CorLoc in VOC 2007. It jointly trains an weak segmentation network with its detection network and refines the detection results by iterations. Compared to its baseline, the improvement is limited.  

\begin{table}[htbp]
	\caption{Comparison with other methods. It demonstrates that our method (OICR+OPG) shows a comparable performance to other state-of-the-art methods. Most listed methods are trained in a fixed training proposal set. Our OPG can be applied to these listed methods to further improve their performances.}
	\begin{center}
		\begin{tabular}{c|c c|c c}
			\hline
			Dataset &\multicolumn{2}{c}{VOC 2007}&\multicolumn{2}{|c}{VOC 2012}\\
			\hline
			Method & mAP   & CorLoc  & mAP  &CorLoc\\
			\hline
			$\rm{TS^2C}$\cite{wei2018ts2c} & 48.0 & 61.0 & 44.4 & 64.0 \\
			WeakRPN\cite{tang2018weakly}    & 50.4 &  68.4 & 45.7& 69.3\\
			CL \cite{wang2018collaborative} & 48.3 & 64.7 &  43.3 & 65.2 \\
			WS-JDS \cite{shen2019cyclic} &52.5 & 68.6& 46.1 & 69.5 \\
			C-MIDN \cite{gao2019c} & 53.6 & 68.7 &  50.3 & 71.2 \\
			C-MIL\cite{wan2019c} & 53.1 & 65.0 & 46.7& 67.4 \\ 
			W-RPN \cite{singh2019you}   &  46.9    &  66.5& 43.2 & 67.5\\
			\hline
			OICR + OPG (ours)  & 50.0 &    65.3 &46.2 & 65.8\\
			\hline				
		\end{tabular}
	\end{center}
	\label{table:other2}
\end{table}

\begin{figure*}[htbp]
	\centering
	{\includegraphics[width=\textwidth]{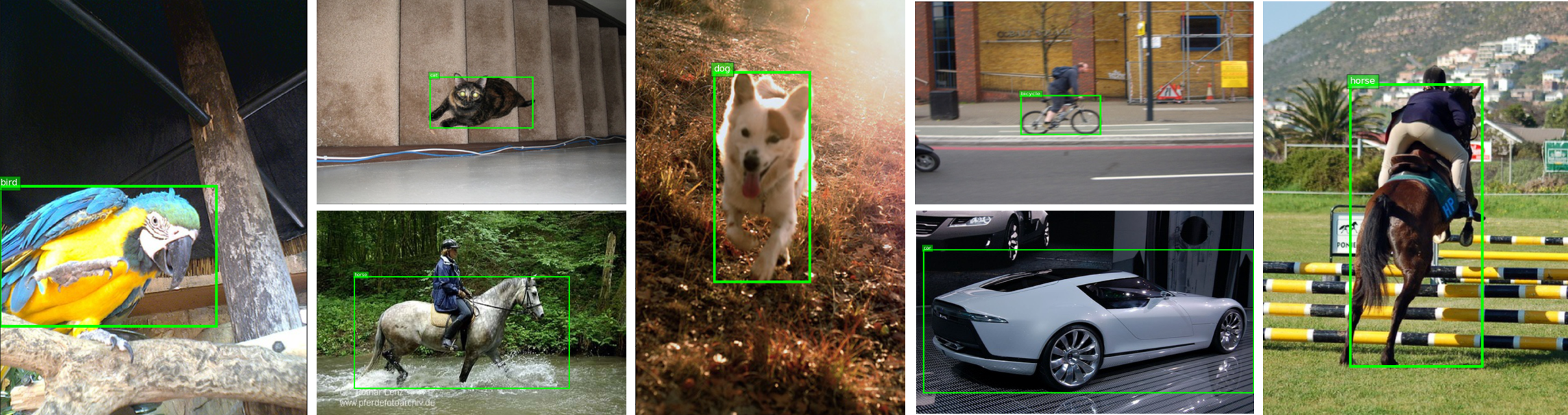}
	}
	\caption{Visualization of our detection results.}
	\label{fig:vis}
\end{figure*}

Our OPG aims to improve the detection accuracy by removing the redundant negative proposals, which is not conflict to most methods in terms of technique. By employing our OPG to other methods, we expect to show a better performance. Some detection results are visualized in Fig. \ref{fig:vis}.    

\section{Conclusion}

In this paper, we have conducted experiments which reveals that the ratio between positive and negative samples affects the detection performance. To solve this problem, we propose an Online Active Proposal Set Generation (OPG) algorithm, which online generates an active training proposal set according to the predictions of the training network. Our OPG can be directly applied to other weakly supervised detection methods, without adding much computation. Experimental results show that our OPG algorithm significantly improves the performance of the baseline and obtains a performance comparable to the state-of-the-art on PASCAL VOC 2007 and 2012 dataset. 

\section*{Acknowledgments}

This work is partly supported by an NTU Start-up Grant (04INS000338C130) and MOE Tier-1 research grants: RG28/18 (S) and RG22/19 (S).

\bibliography{ruibing_paper}

\begin{thebibliography}{10}
\expandafter\ifx\csname url\endcsname\relax
  \def\url#1{\texttt{#1}}\fi
\expandafter\ifx\csname urlprefix\endcsname\relax\def\urlprefix{URL }\fi
\expandafter\ifx\csname href\endcsname\relax
  \def\href#1#2{#2} \def\path#1{#1}\fi

\bibitem{girshick2015fast}
R.~Girshick, Fast r-cnn, in: The IEEE International Conference on Computer
  Vision, 2015.

\bibitem{ren2015faster}
S.~Ren, K.~He, R.~Girshick, J.~Sun, Faster r-cnn: Towards real-time object
  detection with region proposal networks, in: Advances in Neural Information
  Processing Systems, 2015.

\bibitem{redmon2016you}
J.~Redmon, S.~Divvala, R.~Girshick, A.~Farhadi, You only look once: Unified,
  real-time object detection, in: The IEEE Conference on Computer Vision and
  Pattern Recognition, 2016.

\bibitem{liu2016ssd}
W.~Liu, D.~Anguelov, D.~Erhan, C.~Szegedy, S.~Reed, C.-Y. Fu, A.~C. Berg, Ssd:
  Single shot multibox detector, in: European Conference on Computer Vision,
  2016.

\bibitem{li2020weak}
C.~Li, Y.~Huang, H.~Li, X.~Zhang, A weak supervision machine vision detection
  method based on artificial defect simulation, Knowledge-Based Systems 208
  (2020) 106466.

\bibitem{perez2020object}
F.~P{\'e}rez-Hern{\'a}ndez, S.~Tabik, A.~Lamas, R.~Olmos, H.~Fujita,
  F.~Herrera, Object detection binary classifiers methodology based on deep
  learning to identify small objects handled similarly: Application in video
  surveillance, Knowledge-Based Systems (2020) 105590.

\bibitem{wei2020incremental}
X.~Wei, S.~Liu, Y.~Xiang, Z.~Duan, C.~Zhao, Y.~Lu, Incremental learning based
  multi-domain adaptation for object detection, Knowledge-Based Systems 210
  (2020) 106420.

\bibitem{wang2019detection}
Y.~Wang, X.~Luo, L.~Ding, S.~Fu, X.~Wei, Detection based visual tracking with
  convolutional neural network, Knowledge-Based Systems 175 (2019) 62--71.

\bibitem{simonyan2014very}
K.~Simonyan, A.~Zisserman, Very deep convolutional networks for large-scale
  image recognition, arXiv preprint arXiv:1409.1556 (2014).

\bibitem{he2016deep}
K.~He, X.~Zhang, S.~Ren, J.~Sun, Deep residual learning for image recognition,
  in: The IEEE Conference on Computer Vision and Pattern Recognition, 2016.

\bibitem{szegedy2015going}
C.~Szegedy, W.~Liu, Y.~Jia, P.~Sermanet, S.~Reed, D.~Anguelov, D.~Erhan,
  V.~Vanhoucke, A.~Rabinovich, Going deeper with convolutions, in: Proceedings
  of the IEEE conference on computer vision and pattern recognition, 2015, pp.
  1--9.

\bibitem{tang2017multiple}
P.~Tang, X.~Wang, X.~Bai, W.~Liu, Multiple instance detection network with
  online instance classifier refinement, in: Proceedings of the IEEE Conference
  on Computer Vision and Pattern Recognition, 2017, pp. 2843--2851.

\bibitem{tang2018weakly}
P.~Tang, X.~Wang, A.~Wang, Y.~Yan, W.~Liu, J.~Huang, A.~Yuille, Weakly
  supervised region proposal network and object detection, in: Proceedings of
  the European conference on computer vision (ECCV), 2018, pp. 352--368.

\bibitem{wei2018ts2c}
Y.~Wei, Z.~Shen, B.~Cheng, H.~Shi, J.~Xiong, J.~Feng, T.~Huang, Ts2c: Tight box
  mining with surrounding segmentation context for weakly supervised object
  detection, in: Proceedings of the European Conference on Computer Vision
  (ECCV), 2018, pp. 434--450.

\bibitem{wan2019c}
F.~Wan, C.~Liu, W.~Ke, X.~Ji, J.~Jiao, Q.~Ye, C-mil: Continuation multiple
  instance learning for weakly supervised object detection, in: Proceedings of
  the IEEE Conference on Computer Vision and Pattern Recognition, 2019, pp.
  2199--2208.

\bibitem{shen2019cyclic}
Y.~Shen, R.~Ji, Y.~Wang, Y.~Wu, L.~Cao, Cyclic guidance for weakly supervised
  joint detection and segmentation, in: Proceedings of the IEEE Conference on
  Computer Vision and Pattern Recognition, 2019, pp. 697--707.

\bibitem{bilen2016weakly}
H.~Bilen, A.~Vedaldi, Weakly supervised deep detection networks, in:
  Proceedings of the IEEE Conference on Computer Vision and Pattern
  Recognition, 2016, pp. 2846--2854.

\bibitem{zhang2018zigzag}
X.~Zhang, J.~Feng, H.~Xiong, Q.~Tian, Zigzag learning for weakly supervised
  object detection, in: Proceedings of the IEEE Conference on Computer Vision
  and Pattern Recognition, 2018, pp. 4262--4270.

\bibitem{kantorov2016contextlocnet}
V.~Kantorov, M.~Oquab, M.~Cho, I.~Laptev, Contextlocnet: Context-aware deep
  network models for weakly supervised localization, in: European Conference on
  Computer Vision, Springer, 2016, pp. 350--365.

\bibitem{singh2019you}
K.~K. Singh, Y.~J. Lee, You reap what you sow: Using videos to generate high
  precision object proposals for weakly-supervised object detection, in:
  Proceedings of the IEEE Conference on Computer Vision and Pattern
  Recognition, 2019, pp. 9414--9422.

\bibitem{lin2018focal}
T.-Y. Lin, P.~Goyal, R.~Girshick, K.~He, P.~Doll{\'a}r, Focal loss for dense
  object detection, The IEEE International Conference on Computer Vision
  (2017).

\bibitem{shrivastava2016training}
A.~Shrivastava, A.~Gupta, R.~Girshick, Training region-based object detectors
  with online hard example mining, in: Proceedings of the IEEE conference on
  computer vision and pattern recognition, 2016, pp. 761--769.

\bibitem{oquab2015object}
M.~Oquab, L.~Bottou, I.~Laptev, J.~Sivic, Is object localization for
  free?-weakly-supervised learning with convolutional neural networks, in:
  Proceedings of the IEEE Conference on Computer Vision and Pattern
  Recognition, 2015, pp. 685--694.

\bibitem{kosugi2019object}
S.~Kosugi, T.~Yamasaki, K.~Aizawa, Object-aware instance labeling for weakly
  supervised object detection, in: Proceedings of the IEEE International
  Conference on Computer Vision, 2019, pp. 6064--6072.

\bibitem{gao2019c}
Y.~Gao, B.~Liu, N.~Guo, X.~Ye, F.~Wan, H.~You, D.~Fan, C-midn: Coupled multiple
  instance detection network with segmentation guidance for weakly supervised
  object detection, in: Proceedings of the IEEE International Conference on
  Computer Vision, 2019, pp. 9834--9843.

\bibitem{zeng2019wsod2}
Z.~Zeng, B.~Liu, J.~Fu, H.~Chao, L.~Zhang, Wsod2: Learning bottom-up and
  top-down objectness distillation for weakly-supervised object detection, in:
  Proceedings of the IEEE International Conference on Computer Vision, 2019,
  pp. 8292--8300.

\bibitem{pang2019libra}
J.~Pang, K.~Chen, J.~Shi, H.~Feng, W.~Ouyang, D.~Lin, Libra r-cnn: Towards
  balanced learning for object detection, in: Proceedings of the IEEE
  Conference on Computer Vision and Pattern Recognition, 2019, pp. 821--830.

\bibitem{uijlings2013selective}
J.~R. Uijlings, K.~E. Van De~Sande, T.~Gevers, A.~W. Smeulders, Selective
  search for object recognition, International Journal of Computer Vision
  (2013).

\bibitem{zitnick2014edge}
C.~L. Zitnick, P.~Doll{\'a}r, Edge boxes: Locating object proposals from edges,
  in: European conference on computer vision, Springer, 2014, pp. 391--405.

\bibitem{everingham2010pascal}
M.~Everingham, L.~Van~Gool, C.~K. Williams, J.~Winn, A.~Zisserman, The pascal
  visual object classes (voc) challenge, International journal of computer
  vision (2010).

\bibitem{deng2009imagenet}
J.~Deng, W.~Dong, R.~Socher, L.-J. Li, K.~Li, L.~Fei-Fei, Imagenet: A
  large-scale hierarchical image database, in: The IEEE Conference on Computer
  Vision and Pattern Recognition, 2009.

\bibitem{wang2018collaborative}
J.~Wang, J.~Yao, Y.~Zhang, R.~Zhang, Collaborative learning for weakly
  supervised object detection, IJCAI (2018).

\end{thebibliography}

\end{document}